\pdfoutput=1

\documentclass[11pt]{article}

\usepackage[acceptedWithA]{tacl2021v1}

\usepackage{times}
\usepackage{latexsym}

\usepackage[T1]{fontenc}

\usepackage[utf8]{inputenc}

\usepackage{microtype}

\usepackage{inconsolata}

\usepackage{amsmath}
\usepackage{graphicx}
\usepackage{booktabs}
\usepackage{amssymb}
\usepackage{transparent}
\usepackage{multirow}
\usepackage{tikz}
\usepackage[capitalize]{cleveref}
\usepackage{pifont}
\usepackage[super]{nth}
\usepackage[shortlabels]{enumitem}
\usepackage{quoting}
\usepackage{subcaption}
\usepackage{adjustbox}
\usepackage[ruled,vlined]{algorithm2e}
\usepackage{hyperref}
\usepackage{array}
\usepackage{booktabs}
\usepackage{tablefootnote}

\usepackage{xspace,mfirstuc,tabulary}

\newif\iftaclinstructions
\taclinstructionsfalse 
\iftaclinstructions
\renewcommand{\confidential}{}
\renewcommand{\anonsubtext}{(No author info supplied here, for consistency with
TACL-submission anonymization requirements)}
\newcommand{\instr}
\fi

\iftaclpubformat 

\else

\fi

\NewDocumentCommand{\avi}
{ mO{} }{\textcolor{red}{\textsuperscript{\textit{Avi}}\textsf{\textbf{\small[#1]}}}}

\NewDocumentCommand{\sara}
{ mO{} }{\textcolor{blue}{\textsuperscript{\textit{Sara}}\textsf{\textbf{\small[#1]}}}}

\newcommand{\datasetname}{\textsc{CLAPnq}}{}
\newcommand{\datasetmodel}{\textsc{CLAPnq-T5-lg}}{}
\newcommand{\rouge}{RougeL}{}

\usepackage[hang,flushmargin]{footmisc} 

\usepackage{fontawesome5}

\usepackage{xcolor}

\title{\datasetname{}: \underline{C}ohesive \underline{L}ong-form \underline{A}nswers from \underline{P}assages in Natural Questions for RAG systems} 

\author{Sara Rosenthal, Avirup Sil, Radu Florian, Salim Roukos \\
        IBM Research AI \\ \texttt{\{sjrosenthal,avi,raduf,roukos\}@us.ibm.com} \\ 
        }

\begin{document}
\maketitle
\begin{abstract}

Retrieval Augmented Generation (RAG) has become a popular application for large language models. It is preferable that successful RAG systems provide accurate answers that are supported by being grounded in a passage without any hallucinations. While considerable work is required for building a full RAG pipeline, being able to benchmark performance is also necessary.  We present \datasetname{}, a benchmark Long-form Question Answering dataset for the full RAG pipeline. \datasetname{} includes long answers with grounded gold passages from Natural Questions (NQ) and a corpus to perform either retrieval, generation, or the full RAG pipeline. The \datasetname{} answers are \textit{concise}, 3x smaller than the full passage, and \textit{cohesive}, meaning that the answer is composed fluently, often by integrating multiple pieces of the passage that are not contiguous. RAG models must adapt to these properties to be successful at \datasetname{}. We present baseline experiments and analysis for \datasetname{} that highlight areas where there is still significant room for improvement in grounded RAG. \datasetname{} is publicly available at \url{https://github.com/primeqa/clapnq}.
\end{abstract}

\section{Introduction}

Question answering (QA) has been a popular natural language processing task for many years. Large scale research in this area began with the tasks of Machine Reading Comprehension \cite{rajpurkar-etal-2016-squad, 10.1145/3560260MRCdata, mrqa-2021-machine},  and Information Retrieval \cite{manning2008introductionIR, 10.5555/1121636TRECIR, BEIR} and is more recently known as Retrieval Augmented Generation~\cite{lewis2021retrievalaugmented, guu2020realm} which encompasses both tasks. The recent popularity of generative AI with Large Language models (LLM), such as GPT~\cite{brown2020languageGPT}, Llama~\cite{touvron2023llama}, FLAN-T5 \cite{chung2022scalingflant5}, and Mistral~\cite{jiang2023mistral} has shifted the focus to providing long and detailed answers for any user information need. An important challenge for responses produced by an LLM is ensuring that answers are faithful (being grounded in a supporting passage) to ensure that a user can be confident in the response provided to them. 

\begin{figure}[t]
    \centering
    \includegraphics[width=\columnwidth]{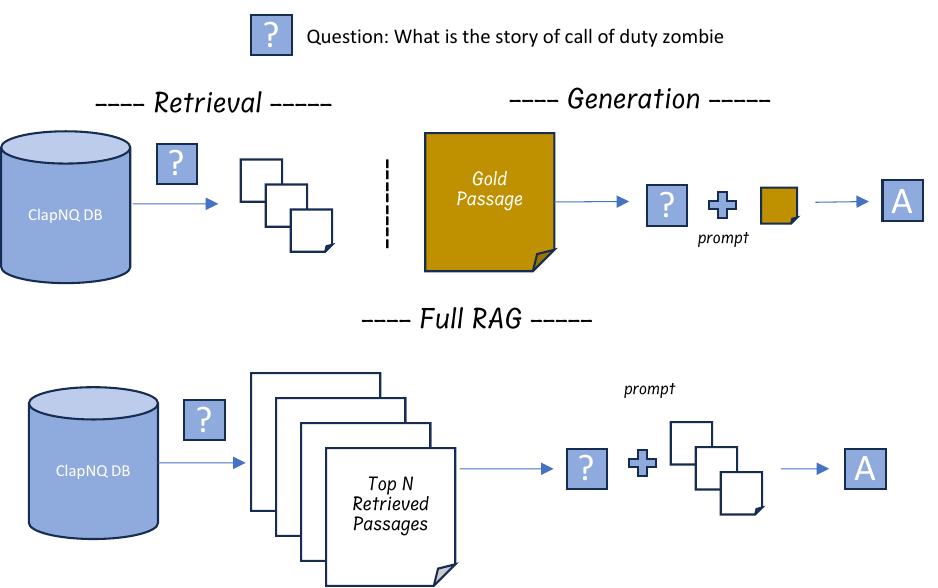}
    \caption{\datasetname{} is designed to test all parts of the RAG pipeline: Retrieval, Generation with gold passages, and the full RAG setup with generation on retrieved passages.}
    \label{fig:rag-pipeline}
\end{figure}

\datasetname{} is a grounded long-form QA benchmark dataset for Retrieval Augmented Generation  of LLMs. The answers are typically long, 2-3 sentences, in contrast to datasets based on machine reading comprehension such as Natural Questions (NQ)~\cite{kwiatkowski-etal-2019-natural} and SQuAD~\cite{rajpurkar-etal-2016-squad, rajpurkar2018knowsquad2} which are just a few words. It is grounded on a single gold passage, in contrast to other long-form question answering (LFQA) datasets such as ELI5~\cite{fan-etal-2019-eli5} where gold passages are not available. It is built from a subset of the highly successful Natural Questions~\cite{kwiatkowski-etal-2019-natural} dataset for extractive QA from Wikipedia documents. The NQ questions are based on users real web search queries. Specifically, we explore the subset of NQ that has long answers (passages) but no short extractive answers. \datasetname{} is suitable for evaluating all parts of Retrieval Augmented Generation (RAG) systems: Retrieval, Generation and the full RAG pipeline (Figure~\ref{fig:rag-pipeline}):

\begin{description}[itemsep=-.5ex,topsep=.5ex,partopsep=1ex,parsep=1ex,leftmargin=*]
\item[Retrieval] Retrieve \textit{N} relevant passages for a question from the indexed \datasetname{} corpus. 
\item[Generation] Generate a response/answer for the prompt which is the concatenation of the question, the gold passage, and the instruction for the model.
\item[RAG] Retrieve \textit{N} passages for the question from the \datasetname{} corpus. Generate a response/answer for the prompt which is the concatenation of the question, \textit{N} passages, and instruction for the model.
\end{description}

\noindent{}It is important to evaluate all RAG scenarios to measure retrieval and generation performance separately, as well as the full pipeline to illustrate how the retrieval performance and noisy passages impacts generation, making it a much more difficult and challenging task.

We present the \datasetname{} dataset of 4946 questions with gold passages for evaluating generation models on grounded LFQA with its corresponding corpus. The answers in \datasetname{} are faithful, concise, complete, and cohesive. An example of a question and grounded answer from \datasetname{} is shown in Table~\ref{tab:example}. We created \datasetname{} with the following properties in order to make it suitable for evaluating generative models:

\begin{description}[itemsep=-.5ex,topsep=.5ex,partopsep=1ex,parsep=1ex,leftmargin=*]
    \item[Faithful] The answer must be grounded in the gold passage. While the answers can be written differently than in the passage, they tend to be highly extractive due to the nature of the dataset creation.
    \item[Concise] The answer must have all the information needed to answer the question but exclude information that is unrelated to the answer. In the original NQ dataset, the entire passage is considered the answer, but this has too much irrelevant information.
    \item[Complete] A short answer (e.g. 2-3 words) commonly found using MRC systems is not sufficient for many types of questions that have a richer information need, require clarity or an explanation. The response must include all information needed to answer the question.
    \item[Cohesive] The response contains the important concepts/facts of the passage assembled together to form a concise and complete answer. In many cases, forming a cohesive answer requires harmonizing multiple non-contiguous pieces of text in a fluent manner.
    \item[Unanswerable] We retain a portion of NQ unanswerable questions that have similar properties to the answerable \datasetname{} questions. This has been largely overlooked by prior LFQA datasets, while expected for real-world RAG applications.
\end{description}

\begin{table}[t]
    \centering
    \small
    \begin{tabular}{p{7.4cm}}
    \toprule
    \textbf{Question:} what is the story of call of duty zombie \\ 
    \midrule
    \textbf{Title:} \textbf{\color{blue}{Call of Duty: Black Ops III}}\\
    \textbf{Passage:} \textbf{\color{blue}{Black Ops III takes place in 2065 , 40 years after the events of Black Ops II , in a world facing upheaval from climate change and new technologies .}} Similar to its predecessors , the story follows a group of black ops soldiers . The game 's campaign is designed to support 4 - player cooperative gameplay , allowing for bigger , more open level design and less corridor shooting . As the player character is cybernetically enhanced , players have access to various special activities . \textbf{\color{blue}{The game also features a standalone Zombies mode , and a `` Nightmares '' mode which replaces all enemies as zombies .} }  \\
    \midrule
    \textbf{Reference Answer}: Call of duty: Black Ops III takes place in 2065 in a world facing upheaval from climate change and new technologies. The game features a standalone Zombies mode, and a `` Nightmares '' mode which replaces all enemies as zombies. \\
    \bottomrule
    \end{tabular}
    \caption{An example of a \datasetname{} answerable question with the reference annotated answer. Sentences in \textbf{\color{blue}{bold}} were selected as relevant parts of the answer. The annotators combined them with modifications to make a cohesive and complete answer.\vspace*{-4mm}}
    \label{tab:example}
\end{table}

\noindent\datasetname{} is the first LFQA benchmark dataset to have grounded gold passages and a full corpus making it suitable for evaluating the full RAG pipeline. Our experiments and results in Section~\ref{sec:experiments} show that LLMs still need considerable work in answering LFQA, remaining faithful to the document, performing the full RAG pipeline, and knowing when a question should not be answered. 

\vspace{1pt}

\noindent Our main contributions are:

\begin{enumerate}[itemsep=-1ex,topsep=.5ex,partopsep=1ex,parsep=1ex,leftmargin=*]
    \item The creation of \datasetname{} with non-consecutive relevant fragments, allowing to test the ability of LLMs to extract just the relevant parts of the passage, while remaining faithful and concise.
    \item A set of baseline experiments with State-of-the-Art (SOTA) models for both retrieval, generation, and the full RAG pipeline.
    \item A human evaluation and discussion to highlight areas where there is room for improvement.
\end{enumerate}

In the rest of this paper we present related work, the dataset creation and details, experiments and results on SOTA retrieval, generative models and the full RAG pipeline. We also present human evaluation, analysis and areas of future research that the \datasetname{} benchmark can be used for advancing RAG research. \datasetname{} is publicly available in a Github repository\footnote{\url{https://github.com/primeqa/clapnq}}. 

\section{Related Work}
\label{sec:related}

\begin{table*}[]
\small
    \centering
    \begin{tabular}{l|l|l|l|l|l|l|l}
    	Dataset & Queries	&	A per Q	&	W in Q	&	W in A	&	S in A	&	 IAA	&	 Unanswerable	\\
     \toprule
AquaMuse Abstractive &	21042	&	1.0	&	9.2	&	106.7	&	3.7	&	-	&	-	\\
AquaMuse Extractive	& 44217	&	1.0	&	9.2	&	106.7	&	3.7	&	-	&	-	\\
ASQA	& 6316	&	1.3	&	10.1	&	80.7	&	3.2	&	0.48	&	-	\\
ELI5	& 1507	&	12.0	&	19.6	&	116.9	&	5.7	&	0.16	&	-	\\
ExpertQA	& 2169	&	1.0	&	21.2	&	174.8	&	6.1	&	-	&	-	\\
TruthfulQA	& 817	&	3.2	&	12.4	&	9.0	&	1.0	&	0.37	&	11	\\
WikiHowQA	& 1188189	&	1.0	&	7.0	&	70.1	&	7.6	&	-	&	-	\\
\datasetname{}-R1	& 12657	&	1.1	&	9.2	&	39.0	&	1.6	&	-	&	-	\\
\datasetname{}	& 4946	&	1.4	&	9.4	&	56.8	&	2.3	&	0.67	&	2493	\\
    \end{tabular}
    \caption{Comparison to existing Long-form QA datasets. Stats are shown for Answers (A), Queries (Q), Words (W), Sentences (S), IAA and Unanswerable. W in A of \datasetname{} is 1/3 of W in Passage (P)=156.}
    \label{tab:lfqastats}
\end{table*}

Natural Questions \cite{kwiatkowski-etal-2019-natural} is a large MRC QA dataset of 323k questions built using Wikipedia documents as the source for natural queries users inputted into Google. Each question was manually annotated given a provided Wikipedia document. There is also an open-retrieval version of NQ, OpenNQ \cite{lee-etal-2019-latent-opennq} where the task is to find the answer to the question via retrieval, but it only focuses on the short extractive answers, and therefore does not include the same set of questions as \datasetname{}. This corpus is also considerably larger than our corpus as we just include the Wikipedia documents used in the \datasetname{} questions. Several datasets have been developed from NQ such as AmbigQA \cite{min-etal-2020-ambigqa}, ASQA \cite{stelmakh-etal-2022-asqa}, AquaMuse \cite{kulkarni2020aquamuse}, AttributedQA~\cite{https://doi.org/10.48550/arxiv.2212.08037AttributedQA}, MoQA~\cite{yen-etal-2023-moqa} and now \datasetname{}.

Several RAG datasets exist for short extractive answers (e.g. ~\cite{lee-etal-2019-latent-opennq, adlakha-etal-2022-topiocqa, https://doi.org/10.48550/arxiv.2212.08037AttributedQA}). MoQA~\cite{yen-etal-2023-moqa} explores answers of varying length but the long answers are full paragraphs as in the original NQ. Current LFQA datasets include AquaMuse \cite{kulkarni2020aquamuse}, ASQA \cite{stelmakh-etal-2022-asqa}, ELI5 \cite{fan-etal-2019-eli5}, ExpertQA~\cite{malaviya2023expertqa}, TruthfulQA~\cite{lin-etal-2022-truthfulqa}, and WikiHowQA~\cite{DBLP:conf/aaai/DengLXCL0S20wikihowqa}. ASQA and ELI5 along with QAMPARI~\cite{amouyal2023qampari} are part of the Automatic LLMs' Citation Evaluation (ALCE) ~\cite{gao-etal-2023-enabling} benchmark. QAMPARI is not LFQA, but rather multiple short extractive answers. We compare all the LFQA datasets to \datasetname{} in Table~\ref{tab:lfqastats}. Most notably, \datasetname{} is the only dataset to include considerable unanswerable questions, manually annotated answers grounded on a single gold passage, and a corpus for the full RAG pipeline. 

The \href{https://www.reddit.com/r/explainlikeimfive/}{Explain Like I'm 5 (ELI5)} dataset consists of questions and responses from the Reddit thread. KILT-ELI5 \cite{petroni-etal-2021-kilt} provides Wikipedia documents that have been retrieved using the questions for benchmarking RAG. However, there are no gold passages and the KILT-ELI5 documents do not necessarily have the answer. The responses written for this sub-Reddit are by subject matter experts (SME) and are 
often not grounded on any text or passage.
Each question is likely to have many responses and they may not all be appropriate or relevant and inter-annotator agreement (IAA) is very low as shown in Table~\ref{tab:lfqastats}.  IAA is measured as the mean \rouge{} F1 score between each pair of annotations for the same question.

TruthfulQA~\cite{lin-etal-2022-truthfulqa} has sets of true and false reference answers and a source that supports the reference answers for each question. It is a very small 
validation dataset as shown in Table~\ref{tab:lfqastats} that was designed to be adversarial (the questions were intentionally picked to be ones that are answered incorrectly) to probe LLMs. The answers are also considerably shorter than the other LFQA datasets.

WikiHowQA~\cite{DBLP:conf/aaai/DengLXCL0S20wikihowqa} is ``How to'' instruction questions from the WikiHow website. For each page, the question is the title and the answer is the context. Only pages that have reference documents are kept. There can be many references for each question. The answers and references are long and have not been manually verified.

ExpertQA~\cite{malaviya2023expertqa} consists of questions that are written by SMEs.
They then use GPT-4 and various retriever setups (e.g. Closed-Book, and BM25) to generate several answers and retrieve relevant documents. The experts then evaluate the answers and evidence and can delete claims and evidence that are false and revise if they want to (it is optional). 
Only one answer was evaluated and revised for each question. Due to the approach of creating the dataset the answers are likely biased by the LLMs~\cite{yu2024large,10.1145/3597307}.

AquaMuse~\cite{kulkarni2020aquamuse} is a summarization dataset using NQ questions that have a long answer (the passage) without a short answer similar to \datasetname{}. However, they use sentence-level matching (by encoding sentences for semantic similarity comparisons) to retrieve up to top 7 documents from Common Crawl while avoiding exact matches as the abstractive dataset. In the extractive version, the sentences in the original long answer are then replaced with the highly semantic similar sentences from the retrieved documents. This means the new summaries are as long as the original passage. The information in the original passage may not be in the retrieved documents. 
 
ASQA~\cite{stelmakh-etal-2022-asqa} has the distinctive characteristic that it uses ambiguous questions built from AmbiqQA \cite{min-etal-2020-ambigqa} which is derived from OpenNQ \cite{lee-etal-2019-latent-opennq}. These ambiguous questions tend to need longer answers to disambiguate the multiple aspects of the question. Each answer is generated from one or more passages that answer a specific instance of the question. The answers in the AmbigQA paper are all short and extractive, but in ASQA the explanation to disambiguate the different answers causes them to be long. ASQA is derived from the subset of NQ that has short answers with additional answers for the ambiguity from AmbigQA. Therefore, the gold passages for the multiple answers are not available for all ASQA questions and some of the evidence may not be part of OpenNQ.
ASQA is perhaps the most similar to \datasetname{}, with the main differences being: 1) The ASQA answer comes from multiple passages while the \datasetname{} answer is contained in one passage. The gold passages are not maintained for ASQA. 2) The ASQA answers are considerably longer due to the ambiguity of the questions, but also indicating they may not be as concise  3) We explore additional types of questions that tend to require a long answer such as boolean questions, conjunctive questions, descriptive questions, and questions requiring an explanation. 4) The IAA computed using \rouge{} for questions that were answered by multiple annotators is much lower than \datasetname{} at 0.48 compared to 0.67.

\begin{table*}[t]
\small
    \centering
    \begin{tabular}{l||l||l|l||l|l}
Split &	No. Questions	& Answerable & NQ	Source	&Unanswerable	& NQ Source \\
\toprule
Train &	3745 &	1954 & Train &1791 & Train \\
Dev	& 600	& 300 & Train	& 300 & Dev \\
Test &	601 &	301	& Train + 67 Dev &	300 & Dev \\ 
\midrule
Total	& 4946  &	2555 & &		2391 & 	\\
\end{tabular}
\caption{Data stats for~\datasetname{}. In addition to providing the number of questions per split we also provide the original source from NQ as we used part of training for the dev and test set.}
\label{tab:clapnqstats}
\end{table*}

For a detailed survey of RAG approaches we direct the reader to the comprehensive RAG survey \cite{gao2024retrievalaugmentedSurvey}. It is worth noting that the benchmarks section in this survey is a short paragraph which refers to two datasets ~\cite{liu2023recallRAG, chen2023benchmarkingRGBRAG} that focus on short extractive answers, attacks and robustness when the passages are purposely adversarial and unfaithful. Furthermore, the datasets questions and responses are created using ChatGPT which likely introduces biases~\cite{yu2024large,10.1145/3597307}. The former~\cite{liu2023recallRAG} does not include retrieval and the latter~\cite{chen2023benchmarkingRGBRAG} has fixed retrieved passages instead of a corpus. We believe that this highlights the need for quality datasets (like \datasetname{}) focusing on faithfulness for the full RAG pipeline.  

Recently, synthetically generated datasets such as Alpaca \cite{alpaca} and Vicuna \cite{vicuna2023}  that include LFQA  have been created using LLMs. These datasets can be very large, containing 50k+ conversations, but they're built to fine-tune LLMs and not applicable as evaluation benchmarks.

\section{Dataset}
\label{sec:dataset}

\datasetname{} is created from the subset of Natural Questions (NQ)~\cite{kwiatkowski-etal-2019-natural} that have a long answer (passage) but no short answer. NQ consists of 323k examples. There are around 30,000 questions that are long answers without short answers excluding tables and lists. To increase the likelihood of longer answers we only explored ones that have more than 5 sentences. Each NQ train example is annotated by one person and each NQ dev example is annotated by 5 people. We only explore dev questions where the majority of the annotators agreed it was a long answer without a short answer. 12,657 training and 384 dev examples met our criteria for annotation.

\subsection{Annotation Task}

\datasetname{} was annotated by 7 skilled in-house annotators paid above minimum wage whose sole jobs are performing Natural Language Processing annotation tasks. The annotation task consisted of two rounds to provide high quality non-consecutive grounded answers to the question. Each task in both rounds took approximately 5 minutes. All annotations were performed on the Appen platform.\footnote{\url{https://www.appen.com/}} The details of each round are described below. 

The main instruction provided to the annotators was: \textit{Given a question and a passage, find the answer to the question in the passage. Check the boxes for the answer sentences and then copy/paste the relevant text into the answer box.} Finally, after creating an answer from the passage they were asked to \textit{look over the question and answer and make sure it makes sense, is a concise answer, and is grammatically correct.} They had to confirm that they checked all of these things before completing the task. A screenshot of the task is provided in Appendix~\ref{app:annotation}, Figure~\ref{fig:annotation_round1}.

After initial training and pilots with calibrating of instructions on around 100 questions, each of the NQ questions without a short answer was annotated by one trained annotator in Round 1. 

In \textbf{Round 1}, the annotators were provided with the question, title, and long answer paragraph from NQ divided into sentences using a sentence tokenizer. The annotators had to select the sentences relevant to the answer and then write a concise answer in their own words with ``copy/pasting'' allowed. The annotators were instructed to write the answer using the selected sentences and that it should make sense, be concise, and grammatically correct. The question could also be skipped.

In \textbf{Round 2} of the annotation, all answers from Round 1 that were made up of two or more selected sentences that were \textit{not consecutive} (meaning there was at least one non-selected sentence between them, see example in Table~\ref{tab:example}) were annotated a second time by a different annotator. These questions were selected as they are more likely to require harmonizing multiple non-contiguous pieces of text. The annotators saw the answer from the first round and could choose to keep the same answer or modify it. Therefore, the second round answers are likely to be of higher quality, however, due to human subjectivity both answers could still be good. In some cases, the round 2 annotator skipped the question and it is also possible that they changed the answer to no longer be non-consecutive.

The final \datasetname{} dataset consists of all answers that have been annotated by more than one person.
We provide the annotations from both rounds if they were different. The IAA using \rouge{} on the different Round 1 and 2 answers is 0.67, indicating the answers are usually similar. The selected sentences, information regarding the round, and whether the answer is not contiguous is included in the dataset.

\subsection{Data Stats}

The \datasetname{} dataset of 4,946 questions consists of both answerable and unanswerable questions as described below. The breakdown of the dataset is shown in Table~\ref{tab:clapnqstats}. We also include the source of the questions within the original NQ dataset. Since NQ does not release the test set we only explored the train and development sets. Only 67 NQ dev questions qualified with the properties of our task so we use them and additional examples from NQ train as our test set. While the questions and passages are publicly available with NQ, the answers we provide are new. \datasetname{} questions have 1-2 reference answers. The questions are short at 9 words and the answers are long at around 57 words which is 1/3 of the average passage length of 156 words (See Table~\ref{tab:lfqastats}). In addition to the official dataset, we will release the round 1 data of 12k questions as training data, referred to as \datasetname{}-R1. Our initial experiments with training using \datasetname{}-R1 did not provide an improvement. We leave further exploration as future work. 

\subsubsection{Answerable}

The answerable data contains the original question and gold passage (P) as well as the relevant sentences (RS) and answers (A) created by the annotators as described in the previous section. The  Precision, Recall (R), and F1 scores for \rouge{}$_{(RS,P)}$ is 100/45/59 and for \rouge{}$_{(A,RS)}$ it is 92/72/79 respectively. The former scores are a sentence retrieval task, the latter a generative task. \rouge{}$_{(A, P)}$ is 94/32/46. The retrieval stage reduces the content by about 2x (R=45) and the generation case reduces another 30\% (R=72) for a total reduction From P to A of approximately 3x (R=32).

\subsubsection{Unanswerable}
A similar amount of unanswerable questions from NQ were extracted to complete the \datasetname{} dataset. In the NQ training set there is only one annotation, in the NQ dev set all 5 annotators must have said it was unanswerable. The unanswerable questions were randomly chosen from examples that had more than 5 sentences in the passage by matching the first word distribution of the answerable questions. For example, in \datasetname{}, \textit{What} and \textit{Where} are the most common question types while \textit{Who} is the most common question type for the NQ short answers. Since NQ does not have a gold passage for unanswerable questions, a random passage is chosen from the provided Wikipedia document. This passage is used in the generation experiments as the "gold" passage to indicate the question is unanswerable with this passage.

\subsection{Retrieval Corpus}
\label{sec:retrievalcorpus}

\begin{table*}[t]
    \centering
    \small
    \begin{tabular}{r|c|c|c|c|c||c|c|c|c|c}
	& \multicolumn{5}{c||}{\textbf{DEV}}			&	\multicolumn{5}{c}{\textbf{TEST}}			\\
 \toprule
 & \multicolumn{4}{c|}{nDCG} & R			&	\multicolumn{4}{c|}{nDCG}	& R		\\
Model	& @1 & @3 & @5 & @10	&	@10	& @1 & @3 & @5 &	@10	&	@10	\\
 \toprule
BM25 & 18 & 30 &	35 &	40 & 67& 20  & 31 &	36 & 40 & 64 \\
all-MiniLM-L6-v2	& 29 & 43 & 48 &	53	&	79	& 30 & 45 & 51 &	55	&	83	\\
BGE-base	&  37 & 54 & 59 &	61	&	85	& 43 & 57 & 63 & 65	&	88	\\
E5-base-v2 & 41 &	57 & 61 &	64	&	87	& 42 &	57 & 61 & 65	&	88	\\
    \end{tabular}
    \caption{Retrieval Results on the answerable questions using nDCG @1, 3, 5, 10 and Recall@10 as metrics on the dev and test sets. We report several nDCG@k to illustrate the impact on the RAG task.}
    \label{tab:retrieval}
\end{table*}

We provide a corpus that can be used to build an index for querying \datasetname{} in a retrieval setting. It is built using the passages\footnote{Very long (> 3000 words) and short passages (<15 words) that are not gold answerable passages were discarded.} from the original Wikipedia NQ documents used in the \datasetname{} dataset including the answerable and unanswerable questions. 

In some cases there were slightly different versions of the same document. We only kept one to avoid duplicate passage retrieval and ensure that all gold passages are present in the corpus\footnote{We confirmed that the gold passage had
very high overlap (RougeL > .90) to the alternative version of the passage or added it as an additional passage for the document (28 times).}. The corpus includes 178,891 passages from 4,293 documents, of which 2,345 passages have questions associated with them across the 4,946 train, dev, and test answerable and unanswerable splits.\footnote{There is usually one gold passage, but 14 questions from the NQ dev set have two gold passages. Both are kept in retrieval, but only the more frequent one has a gold answer.}

\section{Experiments and Results}
\label{sec:experiments}

We present baseline experiments on \datasetname{} for Retrieval, Generation and the full RAG pipeline. An exhaustive implementation of methods and training setups is beyond the scope of this paper; we provide results to illustrate how \datasetname{} performs using common and SOTA approaches.

We report the commonly used retrieval metrics of nDCG@10 and Recall@10 for retrieval. We report several metrics to illustrate generation performance. Each of our metrics correlate with one of the \datasetname{} properties described in the introduction. The first two are the commonly used \rouge{} and Recall (this is the same as Rouge1). \rouge{} can be considered a good approximation for how \textit{cohesive} the answer is as it will give more credit to longer spans. Recall is a good approximation for \textit{completeness}. We also provide \rouge{}$_p$ which is an extractiveness metric that measures how \textit{faithful} the response is. It computes the \rouge{} of the answer to the passage. Since \datasetname{} is extractive, we would expect a good system to have a high \rouge{}$_p$. In addition, we also provide the length (in characters) of the answer. We notice that length is a strong indicator of how well a model performs with answers that are close to the reference length being desirable, it is therefore a good approximating for how \textit{concise} the answer is. Looking at all four of these metrics helps provide a comprehensive picture of model performance. Finally, we also provide the \textit{unanswerable} accuracy. The output is considered unanswerable if its answer string indicates it is unanswerable, e.g. ``I don't know". The unanswerable strings differ per model. 

\subsection{Retrieval}
\label{sec:retrieval}

We present retrieval results on popular public SOTA\footnote{See the Retrieval tab of the MTEB leaderboard: \url{https://huggingface.co/spaces/mteb/leaderboard}} base-size (768 embedding dimension) retrieval dense embedding models E5~\cite{wang2024texte5}, BGE~\cite{chen2024bge}, and allMiniLM\footnote{\url{https://huggingface.co/sentence-transformers/all-MiniLM-L6-v2}} (384 embedding dimension)
in addition to BM25~\cite{robertson2009probabilisticBM25} by ingesting the \datasetname{} corpus described in Section~\ref{sec:retrievalcorpus}. We ran the ingestion and evaluation for the embedding models using sentence transformers from the BEIR repository\footnote{\url{https://github.com/beir-cellar/beir/}} keeping all default parameters, and we used ElasticSearch\footnote{\url{https://www.elastic.co/elasticsearch}} for BM25 with a maximum passage length of 512 tokens. Passages that exceeded the length were divided with an overlap stride of 256. We provide nDCG results for 1, 3 and 5 in addition to 10 to illustrate the potential impact on the full RAG pipeline which we report in Section~\ref{sec:RAG}. The retrieval results are shown on the answerable questions from the ~\datasetname{} dev and test sets in Table~\ref{tab:retrieval}. The E5-Base model performs best with nDCG@10 of 64 on the dev set and E5-base and BGE-base have the same performance of nDCG@10 of 65 on the test set. All these models include NQ as part of their training.

\subsection{Generation}
\label{sec:generation}

\begin{table*}[t]
\setlength{\tabcolsep}{4pt}
\small
    \begin{tabular}{r|r|r|r|r|r|r||r|r|r|r|r}
    & & \multicolumn{5}{c||}{\textbf{DEV}} & \multicolumn{5}{c}{\textbf{TEST}} \\
    \toprule
    & & \multicolumn{4}{c|}{Answerable} & Un- & \multicolumn{4}{c|}{Answerable} & Un- \\
    Model	&	FS	&	\rouge{} &	R & \rouge{}$_p$ & Len &	ans\% & \rouge{} &	R & \rouge{}$_p$ & Len &	ans\%	\\
    \toprule
FLAN-T5-Large		&		- &		18.6		&		11.8		&  7.1	& 33 &	79.9		&	13.8			&		8.5		&	5.0 & 27  & 83.6	\\
FLAN-T5-Large		&		1/0		&		22.0		&		14.6		& 8.8	& 41 &	77.3 &		17.1		&	11.4			&	6.9 & 36 &	82.6		\\	
FLAN-T5-Large		&		1/1	&		20.3	&		13.4		& 8.1	& 38 &	81.7		&	16.3			& 10.4				& 6.1	 & 34  & 85.3	\\
FLAN-T5-XXL		&		- &		22.1		&		15.0		&  10.0	& 45 &	84.0		&	22.0			&		15.6		&	9.7 & 56  & 91.5	\\
FLAN-T5-XXL		&		1/0		&		31.9		&		23.6		& 15.0	& 75 &	78.1 &		28.9		&	21.1			&	14.3 & 76 &	84.9		\\	
FLAN-T5-XXL		&		1/1	&		28.3	&		21.1		& 13.0	& 63 &	84.8		&	24.0			& 17.2				&11.4	 & 63  & 89.2	\\
\midrule
Llama-13B-chat		&		- &		35.5		&		64.3		&	34.0 & 491 &	25.0 &		35.0		&	61.3			&	34.0  & 491  &	27.4	\\	
GPT 4 &		-	&	35.9	&	67.7	& 30.0  & 759 &	18.0	& 33.4		& 65.1		& 30.3  & 797  &	22.2 	\\
Mistral-7B-Instruct &		-	&		39.0		&		56.0		&  29.0	& 384 &	18.6	&		35.4		&	53.4 & 29.2  & 411 & 16.3	 \\
GPT 3.5 &		-		&	39.8	&	58.9	& 30.0 & 444 &	37.0 &	40.3	&	56.3	& 29.9 & 375 &	31.3	\\

\midrule
\datasetmodel{}-200 & - & 41.5 & 51.3 & 42.1 & 272 & 89.7 & 40.5 & 49.2 & 39.0 & 271 & 92.0 \\
\datasetmodel{}  &		-	& 57.2 & 68.3 & 51.0 &  318 & 89.2	&		57.7		&	69.5			&	51.7 & 351  &	86.8 \\
\midrule
Full Passage  &		-	& 49.5 & 97.4 & 100.0 &  912 & 0.0	&		49.2		&	98.7			&	100.0 & 1039  &	0.0 \\
\end{tabular}
\caption{Generation results with the gold passage using \rouge{}, Recall, \rouge{}$_p$, Length and Unanswerable accuracy as metrics. Experiments using pre-trained models, few-shot (1 answerable / 1 unanswerable examples), the fine-tuned model, \datasetmodel{}, and a full passage baseline.}
\label{tab:generation_results}
\end{table*}

The generation task is: \textit{Given a question and the gold relevant passage, generate an answer to the question}. The \datasetname{} dataset is designed to be faithful and concise so the generated response should have these properties. 

We ran generation experiments with three families of models: Encoder-Decoder, Decoder LLMs, and Fine-Tuned Encoder Decoder. We also compare to a full passage baseline. The generation task is sent to the model using a prompt. Most models use an NQ prompt taken from FLAN-T5 \cite{chung2022scalingflant5}. GPT and Llama have prompts based on their model suggestions, all prompts are provided in Appendix~\ref{sec:app_prompts}. In our zero-shot setup the models were provided with the question, context, and prompt. In the 1-shot setup (1/0) the model was provided with the same answerable example from \datasetname{} training and in the 2-shot setup (1/1) the model was also provided with the same unanswerable question for the same passage. The generation results are shown in Table~\ref{tab:generation_results}. A human evaluation and discussion is in Sections~\ref{sec:humaneval} and~\ref{sec:discussion}.

\begin{table*}[t]
\small
\setlength{\tabcolsep}{2.2pt}
\begin{tabular}{l|l|r|r|r|r|r||r|r|r|r|r}
    & & \multicolumn{5}{c||}{\textbf{DEV}} & \multicolumn{5}{c}{\textbf{TEST}} \\
    \toprule
   & & \multicolumn{4}{c|}{Answerable} & Un- & \multicolumn{4}{c|}{Answerable} & Un- \\
    Retriever & Generator	&	\rouge{} &	R & \rouge{}$_p$ & Len &	ans\% & \rouge{} &	R & \rouge{}$_p$ & Len &	ans\%	\\
    \toprule
GOLD & GPT 3.5 &		39.8	&	58.9	& 30.0 & 444 &	37.0 &	40.3	&	56.3	& 29.9 & 375 &	31.3	\\
E5-base-v2 & GPT 3.5 & 34.0 & 52.8 & 30.0 & 459 & 27.3 & 35.0 & 48.9 & 31.4 & 373 & 20.2 \\
\midrule
GOLD & Mistral-7B-Instruct 	&		39.0		&		56.0		&  29.0	& 384 &	18.6	&		35.4		&	53.4 & 29.2  & 411 & 16.3	 \\
E5-base-v2 & Mistral-7B-Instruct & 31.3 & 49.4 & 30.1 & 436 & 11.7 & 29.4 & 47.5 & 29.9 & 463 & 9.3 \\
\midrule
GOLD & \datasetmodel{}	&	57.3 & 68.3 & 51.0 &  317 & 89.5	&		57.8		&	69.5			&	51.7 & 351  &	86.8	\\	
\footnotesize{all-MiniLM-L6v2} & \datasetmodel{} 	&	36.6	& 46.4 & 52.6 & 300	&	49.8 & 37.9 & 48.7 & 52.9 & 323 & 47.0	\\
BGE-base & \datasetmodel{} & 40.7 & 52.3 & 54.2 & 331 & 41.9 & 41.7 & 52.4 & 54.8 & 331 & 44.4 \\
E5-base-v2 & \datasetmodel{} 	&	42.8	&	54.3	& 53.8 & 343 &	40.1 & 41.6 & 51.3 & 55.7 & 321 & 45.9\\
\midrule

E5-base-v2 & E5-\datasetmodel{} & 30.4 & 37.5 & 34.3 & 204 & 82.7  & 26.7 & 32.9 & 33.0 & 195 & 84.6 \\
E5-base-v2 & \footnotesize{E5-G-\datasetmodel{}} &  33.3 & 40.4 & 37.0 & 227 & 78.8 &  34.5 & 41.8 & 38.0 & 236 & 81.0 \\

\end{tabular}
\caption{Full RAG results with top 3 passages on \datasetmodel{} and LLMs using various retrievers. The metrics reported are \rouge{}, Recall, \rouge{}$_p$, Length and Unanswerable accuracy. Each RAG setup can be compared to its GOLD setup where there is no retrieval.}
\label{tab:rag_results}
\end{table*}

\textbf{Encoder Decoder Models.} We use FLAN-T5-Large and FLAN-T5-XXL for zero and few-shot experiments. We chose FLAN-T5 as it has already been trained on the NQ dataset and should therefore already be familiar with the task. The FLAN-T5 models, which are fine-tuned on short extractive tasks, like to provide short answers and therefore have poor Recall. The few-shot experiments outperform the zero-shot experiments, but providing an unanswerable example has a trade-off of improving the unanswerable metrics while reducing the answerable metrics. 

\textbf{Decoder LLMs.} We explored several SOTA Decoder models: LLama, Mistral, GPT 3.5 turbo and GPT 4 turbo. The SOTA LLMs have poor unanswerable performance but better recall. They do not like to say ``I don't know" and almost always provide an answer. This is evident with all models but worst with Mistral and GPT 4. Interestingly, GPT 3.5 performed better than GPT 4, particularly for unanswerable. The LLMs tend to provide answers that are far too long, particularly for GPT 4 at an average of 759 /797 characters, and therefore are not concise. This is apparent from the high Recall but low \rouge{}. The low \rouge{}$_p$ indicates that the answers may not be faithful to the passage.

\textbf{Fine-Tuned Encoder Decoder Model.} We use FLAN-T5-Large for our fine-tuned (FT) experiment, which we call \datasetmodel{} (See implementation details in Appendix~\ref{app:implementation}). \datasetmodel{} has good unanswerable performance and good recall. It is clear that the answers are concise and it learns the appropriate answer length. It is closest to the average length of the reference responses which is 272 dev and 300 test characters. \rouge{} and Recall highlight that the answers are most cohesive and complete and \rouge{$_p$} shows that it learns to extract the answer from the passage, while the other models are considerably less extractive. 

We also explore a smaller training size to help measure whether performance can be improved when a small amount of labeled data is available. This is an important use case because labeling data in a new domain is costly. We call this experiment \datasetmodel{}-200 as it was trained using 200 examples (an equal amount of answerable and unanswerable questions) with 10 random samples and report the average. The \rouge{} and unanswerable metrics are better than the SOTA Decoder LLMs, but worse than training on the full dataset. The model tends to say unanswerable too much.

\textbf{Full Passage Baseline.} We compare to a baseline where the entire passage is taken as the answer. This performs very well in the automated metrics but it is clearly not concise as indicated by the length. The \rouge{} score highlights the difference of the LLMs to \datasetmodel{} which are considerably lower than providing the full passage. The difference between the average length of the generated answers, the reference answer, and the passage length are an indicator of how difficult the extraction task is. The answer must discard two thirds of the passage to be appropriately concise.

\subsection{Full RAG Pipeline}
\label{sec:RAG}

In our full RAG pipeline experiments we retrieve the top passages using the best performing retrieval model, E5-base-v2, and then perform generation on the same prompts as in Section~\ref{sec:generation}, however instead of the gold passage, the top retrieved passages are included in the prompt. It is possible that the gold passage will not be in the top \textit{N} passages making the question unanswerable based on retrieval. The RAG task is far more difficult than the GOLD generation task as the model needs to learn which passages are irrelevant to the question. We experimented with including the top 3 and top 5 passages in the prompt. Based on the retrieval results in Table~\ref{tab:retrieval}, 5 documents has a 4 point improvement over 3 documents. However, in our experiments including 5 passages in the prompt increased the noise and did not provide an improvement.

In the RAG experiments we explored each dense retriever with \datasetmodel{}, and the best retriever on the dev set, E5 Base, with the best performing generation models: GPT 3.5, Mistral-7b-Instruct and \datasetmodel{}. Results are shown in Table~\ref{tab:rag_results} and we compare against the best GOLD generation baselines for each model from Table~\ref{tab:generation_results} to show the gap for RAG. GOLD can be considered as an upper bound as we would not expect the retriever to perform better than having only the grounded passage for the automated metrics. In all cases performance drops considerably for \datasetmodel{} with a very large drop in \% unanswerable. Performance is also reduced for zero-shot GPT 3.5 and Mistral but not as much as~\datasetmodel{}.  A human evaluation and discussion that compares RAG to Gold is in Sections~\ref{sec:humaneval} and~\ref{sec:discussion}.

We also explored two fine-tuned models that incorporated RAG during training. They follow the same approach as \datasetmodel{}, but instead of the gold passage, the top 3 retrieval passages are included during training. In the second version, \textsc{E5-G-\datasetmodel{}} we ensure the gold passage is kept in the top 3 passages during training, at a randomly chosen position, even if it was not originally included. 
These models perform better on the unanswerable questions than ~\datasetmodel{} but much worse on the answerable questions. The \rouge{} score of \textsc{E5-G-\datasetmodel{}} (51.6/52.1) on the answerable questions that were answered is better than \datasetmodel{} (46.7/44.5) for the dev and test sets, but only a little more than half the answerable questions were answered. We suspect the discrepancy for unanswerables between the GOLD experiment and RAG trained models (89.5 vs 82.7/78.8) is because many questions are no longer unanswerable in the RAG setting as we will show in the human eval. We leave further experimentation on optimizing these models as future work. 

\section{Human Evaluation}
\label{sec:humaneval}

In addition to reporting automated metrics we also performed a human evaluation on the GOLD and RAG setups to explore how appropriate and faithful users think the responses are as used in the literature~\cite{es2023ragas}. For each question and answer, we asked three annotators to indicate on a scale of 1 (No) - 4 (Yes) whether the answer looks appropriate (i.e. looks correct or answer relevance) and whether it is faithful to the passage. These metrics are only measured for the answerable questions. During the RAG evaluation we also asked the annotators to select which of the top 3 retrieved passages were relevant to the answering the question. If a question was marked faithful, we asked the annotators to select which passages were relevant to the answer. Finally, they performed a pair-wise comparison of the answers to indicate preference to compute win-rate. Ties were acceptable but they were asked to do so sparingly.
The answers were shown to the annotators randomly and they did not know which model produced the answer. Instructions and a task screenshot 
are in Appendix~\ref{app:annotation}.

The human evaluation was for the GOLD and RAG setups. 40 answerable and 10 unanswerable questions, with an equal amount of questions were randomly sampled from both the dev and test sets being included for each setup. The annotators that performed this task are the same annotators that worked on creating the dataset, however these annotations were done at a later time period. We compare \datasetmodel{}, GPT 3.5 (The best performing decoder LLM), and the reference answer. 
The evaluation is shown in Table~\ref{tab:human_eval}.

\begin{table}[t]
    \setlength{\tabcolsep}{2.7pt}
        \small
    \begin{tabular}{c|r|c|c|c|c}
       & Model & Faithful & Approp & F+A & \footnotesize{Win-Rate} \\
       \toprule
       \multirow{3}{*}{\rotatebox{-90}{Gold}} & 
       \footnotesize{\datasetmodel{}}  & 3.7 & 3.7 & 3.7 & 66\% \\
       & GPT 3.5 & 3.3 & 3.6 & 3.4 & 34\% \\
       & Reference & 3.9 & 3.8 & 3.8 & 57\% \\
       \midrule
       \multirow{3}{*}{\rotatebox{-90}{RAG}} & 
       \footnotesize{\datasetmodel{}}  & 3.8 & 3.2 & 3.4 & 42\% \\
       & GPT 3.5 & 3.0 & 3.6 & 3.2 & 35\%\\
       & Reference & 3.0 & 3.5 & 3.0 & 33\% \\
       \bottomrule
    \end{tabular}
    \caption{Human Evaluation metrics on Faithful (F) and Appropriate (A) on a 4-point scale and win-rate. F+A is the harmonic mean of F and A.}
    \label{tab:human_eval}
\end{table}

\begin{table*}[t]
\centering
\setlength{\tabcolsep}{3pt}
    \small
    \begin{tabular}{l|r|r|r|r||r|r|r|r}
    & \multicolumn{4}{c||}{\textbf{\datasetname{}}} & \multicolumn{4}{c}{\textbf{ASQA}} \\
    \toprule
	 & \rouge{}	&	R & \rouge{}$_p$ & Len & \rouge{}	&	R & \rouge{}$_p$ & Len	\\
\toprule
ASQA-T5-\textsc{lg}	& 51.3 & 77.5 & 67.7 &  551	&		54.9		&	69.5			&	89.5 & 542 \\
\datasetmodel{}-\textsc{ans}	& 62.8 & 74.1 & 54.8 &  364 	&		44.9		&	42.1			&	72.4 & 217 \\

CLAP\textsc{nq}+ASQA-T5-\textsc{lg}-\textsc{ans}	& 61.3 & 74.2 & 57.0 &  381	&		55.9		&	68.2			&	88.3 & 499  \\
\bottomrule
\end{tabular}
\caption{A comparison of ASQA dev (there is no ASQA test) and \datasetname{} test using models trained on each dataset and the two datasets combined. Since ASQA does not have unanswerable questions, all models and results are on answerable questions only.}
\label{tab:asqa}
\end{table*}

\begin{table}[t]
    \setlength{\tabcolsep}{2.6pt}
        \small
    \begin{tabular}{c|r|c|c|c|c}
       & Model & Faithful & Approp & F+A & \footnotesize{Win-Rate} \\
       \toprule
       \multirow{3}{*}{\rotatebox{-90}{\datasetname{}}} & 
       \footnotesize{ASQA}  & 3.9 & 2.7 & 3.4 & 38\% \\
       & \footnotesize{\datasetname{}} & 4.0 & 3.1  & 3.5 & 68\% \\
        &  \footnotesize{\datasetname{}+ASQA} & 3.9 & 3.2 & 3.5 & 65\% \\
       & Reference & 3.9 & 3.7 & 3.7 & 67\% \\
       \midrule
       \multirow{3}{*}{\rotatebox{-90}{ASQA}} & 
       \footnotesize{ASQA}  & 3.7 & 3.1  & 3.3 & 63\% \\
       & \footnotesize{\datasetname{}} & 3.8 & 3.2  & 3.4 & 65\% \\
        &  \footnotesize{\datasetname{}+ASQA} & 3.7 & 3.1  & 3.2 & 61\% \\
       & Reference & 3.2 & 3.3  & 3.2 & 40\% \\
       \bottomrule
    \end{tabular}
    \caption{Human Evaluation metrics comparing ASQA and \datasetname{} on Faithful (F) and Appropriate (A) on a 4-point scale and win-rate. F+A is the harmonic mean of F and A. All models are T5-\textsc{lg} fine-tuned models trained on answerable data only.}
    \label{tab:human_eval_asqa}
\end{table}

In the GOLD setup, agreement was high for appropriateness (73\%), faithfulness (88\%), and win-rate (86\%). The annotators preferred the \datasetmodel{} answers the most and GPT 3.5 answers the least. We investigated several examples where the \datasetmodel{} answers were preferred to the reference answer and both answers were good but the annotators preferred the direct copying by \datasetmodel{}. The reference and \datasetmodel{} answers were highly faithful and appropriate but GPT 3.5 was less faithful. This highlights the importance of being faithful to the passage as an answer can look correct but not be grounded in the passage which may indicate factually incorrect answers. The human evaluation shows that a model can successfully learn to generate faithful and appropriate responses, but the SOTA LLM models don't perform as well on this task.

In the RAG setup, agreement was very high for faithfulness (91\%) and win-rate (90\%) but much lower for appropriateness (68\%). The annotators preferred the \datasetmodel{} answers the most with little difference in preference between the reference and GPT 3.5 answers. The \datasetmodel{} answers were very faithful while GPT 3.5 and the reference were less faithful. The GPT 3.5 and reference answers were more appropriate while \datasetmodel{} was least appropriate. The changes from the GOLD setup highlight the importance of evaluating the RAG pipeline. The reference answers may not be in the retrieved passages even though they are correct. However, being faithful to the passages can provide an inappropriate answer if the retrieved passages are not relevant to the question. According to two or more annotators, 26/40 answerable questions had multiple relevant passages and 4/40 had no relevant passages. 38, 39 and 32 of \datasetmodel{}, GPT 3.5 and reference responses were considered faithful to one or more passages.  50\% of the unanswerable questions had relevant passages.

\section{Question Style Impact}

We explore the differences between \datasetname{} and ASQA by evaluating how they compare to each other. As described in Section~\ref{sec:related}, the ASQA dataset is most similar to \datasetname{} with a few key differences: The ASQA dataset was also created from a subset of NQ, but the questions come from the short extractive answers in NQ so there is no overlap with \datasetname{}. The ASQA questions are ambiguous and several passages are needed to generate the target answer while \datasetname{} questions are of different types and one passage is needed to generate the target answer. The answers in ASQA tend to be considerably longer on average, at 492 characters compared to 318 characters for \datasetname{}. 

We compare the datasets by running generation experiments in the GOLD setup and a small human evaluation. Since ASQA only provides a dev set, we consider that to be the test set and only compare it to the \datasetname{} test set. Our human evaluation experiment was completed on 20 random questions from each dataset.

We fine tuned an encoder-decoder model for each dataset, as well as a model on both datasets for generation experiments using Flan-T5-\textsc{lg}. Since ASQA only has answerable questions, we only explore the answerable questions for both datasets and train a \datasetmodel{}-\textsc{ans} model on the \datasetname{} answerable subset. (Implementation details are described in Appendix \ref{app:implementation}).

ASQA does not provide all of the context paragraphs needed to generate the answer. They provide the gold passages from the original NQ but only supply ``knowledge'' which is small snippets of text from any Wikipedia document that the annotators felt was needed to make the answer. In our experiments we mimic the oracle setting described in the ASQA paper~\cite{stelmakh-etal-2022-asqa}\footnote{In the paper they only use the longer answer as the reference, while we compare to both gold answers as in our \datasetname{} experiments} by including the context passages and the knowledge as the input context. 

The generation results are shown in Table~\ref{tab:asqa} and the human evaluation results are shown in Table~\ref{tab:human_eval_asqa}. The generation experiments show that each model performs significantly better on its own dataset while the model fine-tuned on both datasets does well on both. The human evaluation shows that ASQA clearly does not perform well on \datasetname{} while the other models perform as well as the reference data (the ASQA answers are faithful but not appropriate). An investigation into some of these examples shows that they had extra information that was not needed to answer the question. On the other hand the human evaluation shows that \datasetname{} does perform well on ASQA but the reference data has a low win-rate (the reference answers are appropriate but not faithful). An investigation into some of these examples showed that there was missing evidence for parts of the answer which can encourage hallucination. We also looked at some cases where \datasetname{} did not do well and found that sometimes it only gave one answer instead of the two answers that needed to be distinguished due to ambiguity. These experiments show that the datasets compliment each other and that \datasetname{} is a more faithful and concise dataset.

\section{Discussion}
\label{sec:discussion}

In this section we describe some challenges we've encountered. We describe them here and provide examples in Appendix~\ref{app:examples}.

\textbf{Unanswerable Questions}: While it is unlikely that the unanswerable questions have an answer in the randomly picked passage, we find that in some cases, there is actually an answer (Appendix~\ref{app:examples}, Table~\ref{tab:answerable_unanswerable}). There are other cases where the answer to an unanswerable question may appear correct when looking at the passage, but the passage may not be relevant (Appendix~\ref{app:examples}, Table~\ref{tab:difficult_passage}).

\textbf{Generation:} GPT 3.5 and Mistral will have answers that are correct but not faithful to the passage (Appendix~\ref{app:examples}, Table~\ref{tab:partially_faithful}, \ref{tab:correct_not_faithful}). 
Since the prompts request that the answer use the passage, such an answer should not be provided, or the response should explain that the answer was found elsewhere. In many cases GPT 3.5 and Mistral give an answer that is considerably longer than \datasetmodel{} and the reference (Appendix~\ref{app:examples}, Table \ref{tab:gpt_too_long}).
The recall is high, but the answer is not concise and has extra irrelevant information. During the human evaluation the annotators tend to prefer the concise answers and will often mark long answers as less appropriate.

\textbf{RAG:} The answers can change considerably due to the multiple passages in RAG compared to GOLD (Appendix~\ref{app:examples}, Table~\ref{tab:gold_in_rag}, \ref{tab:gold_for_RAG}, \ref{tab:gold_not_in_rag}). 
In the RAG setting the automated metrics are much lower than the GOLD setting. However, the answers may be good but just have different information which was found only in the provided passages (Appendix~\ref{app:examples}, Table~\ref{tab:gold_in_rag}).
If irrelevant passages are retrieved, (Appendix~\ref{app:examples}, Table~\ref{tab:bad_retrieval}), 
the reference answer will have low extractiveness, but the other answers may still be incorrect while being grounded which is difficult to identify without human evaluation.

\section{Future Directions}

The automated evaluation, human evaluation and discussion highlight several areas of future directions: 1) Unanswerable Questions: Many of the LLMs struggle with the unanswerable questions and often try to provide an answer. 2) Concise Answers: Many of the LLMs like to provide very long answers that are not concise, which is not preferred by humans. 3) Irrelevant Retrieval: The models will try to answer RAG questions even when the passages are irrelevant, either by being unfaithful or incorrect. 4) Multiple correct answers:  It is harder to evaluate RAG correctly because the answers could be correct but different than the gold. 5) Dataset Enhancements: We hope to add more grounded reference answers, a multilingual version, and other domains.

\section{Conclusion}

We have presented~\datasetname{}, a new benchmark dataset for evaluating the \textit{full} RAG pipeline.
\datasetname{} has the properties of being concise, complete, cohesive, faithful to the passage and unanswerable questions. 
A FT model can perform well when the correct passages are provided during retrieval, while SOTA LLMs are behind in faithfulness, conciseness and unanswerability. Finally, we've provided a human evaluation, discussion, and specific areas of future improvements. \datasetname{} is publicly available at \url{https://github.com/primeqa/clapnq}.

\section*{Ethics Statement}

\subsection*{Limitations}

As with any manually annotated dataset, there are likely to be some incorrect and unclear answers. We did out best to mitigate this as described in Section~\ref{sec:dataset}. We believe in general, that the dataset quality is strong and can be used as is as a benchmark for RAG. \datasetname{} is built from Natural Questions \cite{kwiatkowski-etal-2019-natural}, therefore any limitations in Natural Questions and Wikipedia may also be present in \datasetname{}.

\subsection*{Intended Use}

\datasetname{} and \datasetmodel{} are intended to be used to advance research in RAG. \datasetname{} is being released with an Apache 2.0 license. We do not approve of any adversarial or harmful uses of our work.

\subsection*{Biases}

NQ train and dev have been included in training of most, if not all, LLMs which may lead to biases, particularly since \datasetname{} dev is part of NQ train. However, all models have this same advantage. While the questions and passages have been seen by all models the \datasetname{} answers are new and remain hidden. Any biases in NQ and Wikipedia may also be present in \datasetname{}.

\section*{Acknowledgements}

We would like to thank our annotators for their high quality work creating and evaluating this dataset: Mohamed Nasr, Joekie Gurski, Hee Dong Lee, Roxana Passaro, Marina Variano, and Chie Ugumori. We thank Arafat Sultan for initial annotations and analysis on the pilot task. We would like to thank our TACL Action Editor, Jimmy Lin, and the anonymous reviewers for their helpful feedback.

\bibliography{custom}
\bibliographystyle{acl_natbib}

\clearpage

\appendix

\begin{figure*}[t]
    \centering
    \includegraphics[width=.8\linewidth]{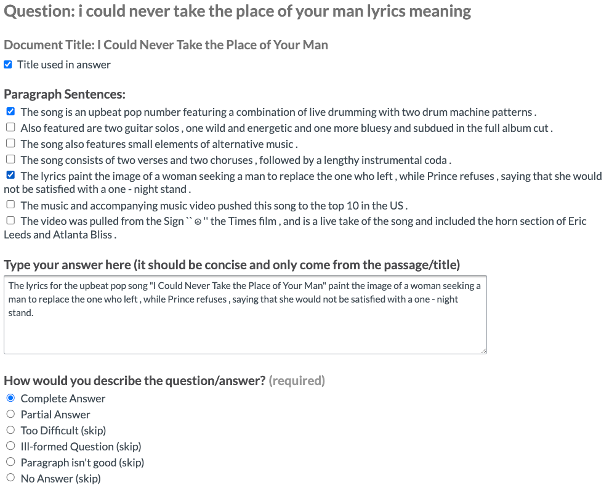}
    \caption{The Round 1 annotation task for \datasetname{}. The annotator had to select the title/sentences needed to answer the question, and then provide a concise answer.}
    \label{fig:annotation_round1}
\end{figure*}

\begin{figure*}[t]
    \centering
    \includegraphics[width=.8\linewidth]{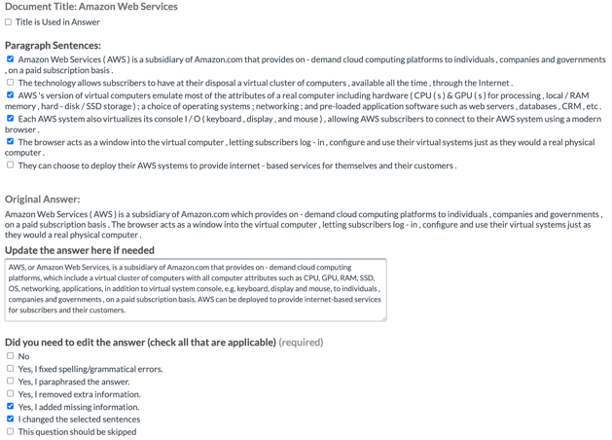}
    \caption{The Round 2 annotation task for \datasetname{}. The annotator had to verify and update the answer provided in Round 1 if needed. They also had to provide how they edited the answer.}
    \label{fig:annotation_round2}
\end{figure*}

\begin{figure*}[t]
    \centering
    \includegraphics[width=0.8\linewidth]{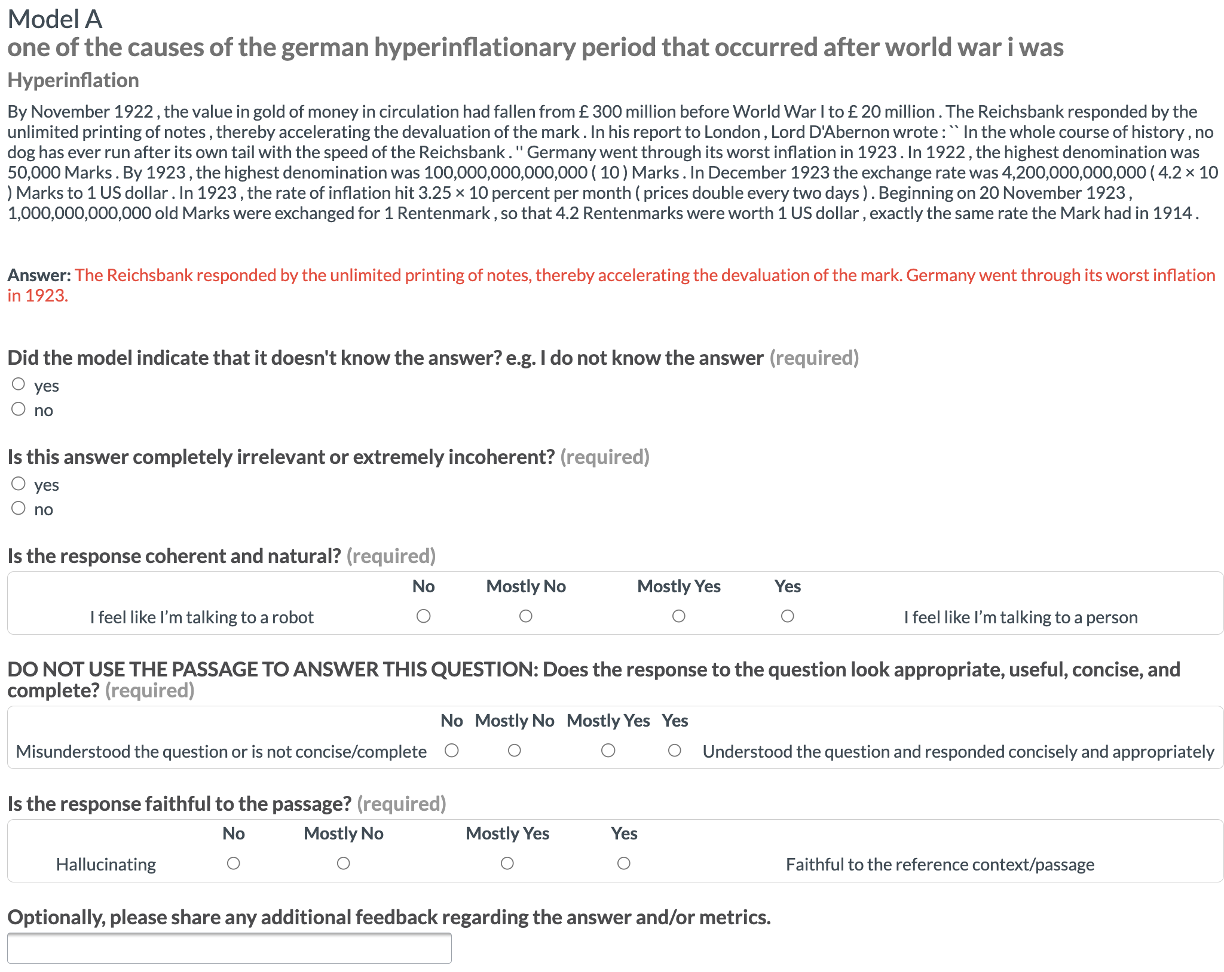}
    \caption{The human evaluation task used to compare the model answers in random order. The individual questions per answer are shown here for one model.}
    \label{fig:human_eval}
\end{figure*}

\begin{figure*}[t]
    \centering
    \includegraphics[width=0.8\linewidth]{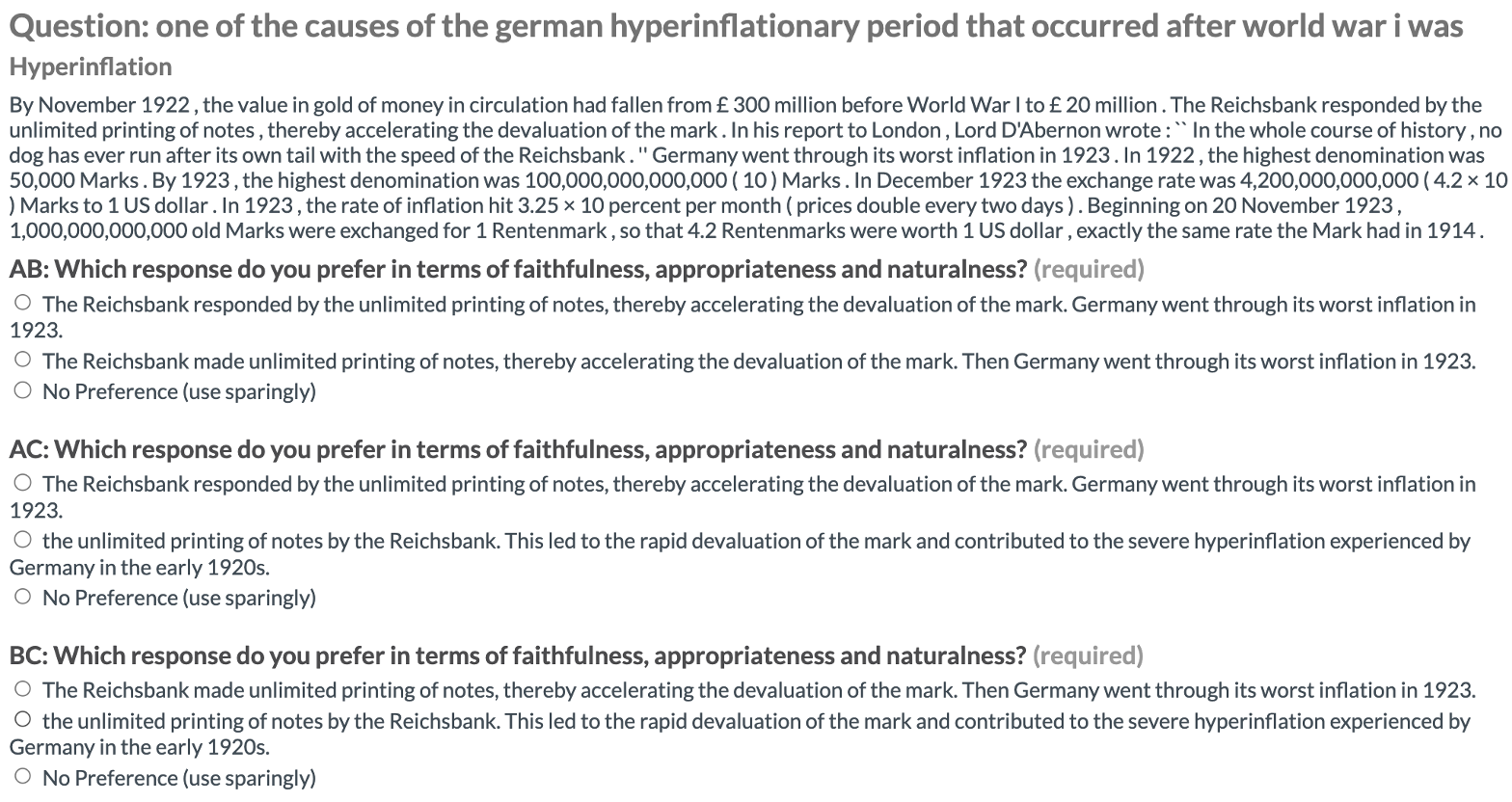}
    \caption{The human evaluation task used to compare the model answers in random order. The head-to-head comparison for win-rate is shown here.}
    \label{fig:win_rate}
\end{figure*}

\section{Annotation Tasks}
\label{app:annotation}


All annotation tasks were performed using Appen. They are described in Section~\ref{sec:dataset} and ~\ref{sec:humaneval} of the main paper. We provide screenshots and further instructions below.

\subsection{Dataset Creation}

The \datasetname{} dataset was created in two rounds. A screenshot of round 1 is shown in Figure~\ref{fig:annotation_round1} and Figure~\ref{fig:annotation_round2}. 
A small handful of the questions (1 in train, and 9 in dev) are high-quality annotations from the initial pilot rounds. These examples have several reference answers.

\subsection{Human Evaluation}

The human evaluation was performed a portion of the dev and test sets. Human eval on the GOLD generation task is shown in Figure~\ref{fig:human_eval} and ~\ref{fig:win_rate}. 
The RAG version had two additional questions regarding passage relevance as described in Section~\ref{sec:humaneval}. We plan on releasing the human evaluation annotations as part of the dataset release. The general instructions to the annotator were as follows:  \textit{In this task, you will review the same question and passage and, for each one, rate the quality of the answer to the question. On each page, you will see 3 different answers to the same question. Read the question and passage and answer how well you are confident in the question, passage, and know the correct answer. 
For each model answer, (given the same context and passage): The answer to the model is in red. Please make your judgements on this red answer span. indicate if the answer is an ``I don't know'' or if the answer is completely incoherent. For each model response, answer the following questions on a scale of 1-4: 1) DO NOT USE THE PASSAGE TO ANSWER THIS QUESTION: Does the response to the question look appropriate, useful, concise, and complete? 2) Is the response faithful to the passage? Evaluate each metric independently. Finally, also perform a head to head comparison of the model responses by answering the following question for every pair of answers:  Which response do you prefer in terms of faithfulness, appropriateness and naturalness?}

 The win-rate is computed per model per question for all annotators. If there are three models and three annotators being compared a model can win up to six times per question (or ($\#$ models - 1) $\times \#$ annotators). The score per model per question is computed as wins/6 which is then averaged over all questions for the final model score.

\section{Prompts}
\label{sec:app_prompts}

\noindent The Flan-T5~\cite{chung2022scalingflant5} prompt which was used for most models is: \{title\}: \{passage\} Please answer a question about this article. If the question is unanswerable, say ``unanswerable''. user: \{question\}, answer:

\noindent The GPT Prompt is based on chat completion from OpenAI\footnote{\url{ https://learn.microsoft.com/en-us/azure/ai-services/openai/reference}}: \{`role': `system', 'content': ``Generate next agent response, given the following document(s). If you cannot base your answer on the document, please state that you do not have an answer.'\}, \{`role': `system', `content': ``[title]: \{title\} [document]:  \{passage\}, \{`role': `user', `content': {question\}'\}

\noindent The Llama Prompt is the default Llama 2 prompt~\cite{touvron2023llama}: $<$s$>$[INST] $<<$SYS$>>$ You are a helpful, respectful and honest assistant. Always answer as helpfully as possible, while being safe. Your answers should not include any harmful, unethical, racist, sexist, toxic, dangerous, or illegal content. Please ensure that your responses are socially unbiased and positive in nature. If a question does not make any sense, or is not factually coherent, explain why instead of answering something not correct. If you don't know the answer to a question, please don't share false information. $<<$/SYS$>>$  [document]: \{title\} \{passage\}. [conversation]: {question\},. Answer with no more than 150 words. If you cannot base your answer on the above document(s), please state that you do not have an answer. [/INST]


\section{Implementation Details}
\label{app:implementation}

We used HuggingFace transformers\footnote{\url{https://huggingface.co/docs/transformers/model_doc/flan-t5}} for all training experiments. We experimented with several variations and our final \datasetmodel{} model was trained for 6 epochs, checking after each epoch against the development set to keep the best model. \datasetmodel{} has a learning rate of $1e-4$, batch size of 32, max input length of 412 tokens with an output length of 100 tokens. We kept the T5 context length during training to keep within GPU constraints and improve training speed. We cut off the end of the 368 passages  (10\% of the training data) that did not fit in the context. The prompts were not truncated during evaluation. 

In the small sample size experiments, we accommodate for the smaller size by increasing the learning rate to $1e-3$. In the RAG \datasetmodel{} experiments, the context is considerably longer so we increase the context size to 1024. This maintains a similar amount of data that needs to be cut off as in the other experiments at 317 and 346 for the E5-\datasetmodel{ }and E5-G-\datasetmodel{} models (<10\%). We accommodate for these experiments with the longer context size by using a batch size of 8 and 10 epochs. In the experiments comparing  \datasetname{} and ASQA we also used the longer context, epochs, and smaller batch size to accomodate for the longer length of ASQA questions and answers.  


\section{Examples}
\label{app:examples}

We provide several examples of output comparing the various LLMs as described in Section~\ref{sec:discussion}. See Tables~\ref{tab:answerable_unanswerable}-\ref{tab:bad_retrieval}.

\begin{table*}
    \small
    \begin{tabular}{p{15.5cm}}
    \toprule
     \textbf{Question:} who are the characters in the black cat \\
     \toprule
     \textbf{Passage:}  Sephiria Arks ( Sefiria Ākusu ) , known as Number I , is the commander of the Numbers and is trusted enough by Chronos to be given total control over killing Creed . Sephiria was one of the few members who were `` born for Chronos '' , as she stated that when she was an infant she underwent surgery to enhance her recovery time from injuries . Sven once commented that her eyes allowed her to stare into the recesses of one 's soul . This empathy is her greatest strength , which allows her to utilize her powers of manipulation to their utmost extent . She will do anything to finish her missions and will sacrifice whatever it takes in order to succeed in meeting an objective . However , she does show some inner conflict in her manipulation , even questioning Belze about her actions . She controls all of the Chronos Numbers ' actions as well as many of the independent operatives in the hunt for Creed with little interference from the Elders .\\
     \toprule
     \multicolumn{1}{|c|}{\textbf{Responses}} \\
     \toprule
     \textbf{GPT 3.5} \\
     \midrule
    In the Black Cat series, there are several characters. One of the main characters is Sephiria Arks, also known as Number I. She is the commander of the Numbers and has complete control over killing Creed. Sephiria underwent surgery as an infant to enhance her recovery time from injuries and has the ability to stare into the recesses of one's soul. She is highly skilled in manipulation and will do whatever it takes to succeed in her missions. Sephiria controls the actions of the Chronos Numbers and many independent operatives in the hunt for Creed.	 \\
    \toprule
    \textbf{\datasetmodel{}} \\
    \midrule
    Sephiria Arks, known as Number I, is the commander of the Numbers and is trusted enough by Chronos to be given total control over killing Creed. Sven once commented that her eyes allowed her to stare into the recesses of one's soul. This empathy is her greatest strength, which allows her to utilize her powers of manipulation to their utmost extent.	\\
    \toprule
    \textbf{Reference} \\
    \midrule
    No answer	\\
    \bottomrule 
    \end{tabular}
    \caption{This is an Unanswerable question that is actually partially answerable from the passage. It only mentions one character.}
    \label{tab:answerable_unanswerable}
\end{table*}

\begin{table*}
\small
    \begin{tabular}{p{15.5cm}}
    \toprule
     \textbf{Question:} what type of volcano is one tree hill \\
     \toprule
     \textbf{Passage:} Maungakiekie One Tree Hill Maungakiekie / One Tree Hill after the removal of the `` one tree '' . Highest point Elevation 182 m ( 597 ft ) Coordinates 36 \textdegree 54 ' 0 '' S 174 \textdegree 46 ' 59 '' E $\cdot$ / $\cdot$ 36.90000 \textdegree S 174.78306 \textdegree E $\cdot$ / - 36.90000 ; 174.78306 Coordinates : 36 \textdegree 54 ' 0 '' S 174 \textdegree 46 ' 59 '' E $\cdot$ / $\cdot$ 36.90000 \textdegree S 174.78306 \textdegree E $\cdot$ / - 36.90000 ; 174.78306 Geography Location North Island , New Zealand Geology Volcanic arc / belt Auckland volcanic field Location \\
     \toprule
     \multicolumn{1}{|c|}{\textbf{Responses}} \\
     \toprule
     \textbf{GPT 3.5} \\
     \midrule
    One Tree Hill is part of the Auckland volcanic field, which is a volcanic arc/belt.	 \\
    \toprule
    \textbf{\datasetmodel{}} \\
    \midrule
    No answer	\\
    \toprule
    \textbf{Reference} \\
    \midrule
    No answer	\\
    \bottomrule 
    \end{tabular}
    \caption{This is a difficult example which confused most annotators during the human evaluation. It appears from the passage that the GPT answer may be correct but the answer is “scoria cone volcano” which is not in the passage.}
    \label{tab:difficult_passage}
\end{table*}
	
\begin{table*}
\small
    \begin{tabular}{p{15.5cm}}
    \toprule
     \textbf{Question:} who is given credit for inventing the printing press \\
     \toprule
     \textbf{Passage:} The world 's first movable type printing press technology for printing paper books was made of porcelain materials and was invented around AD 1040 in China during the Northern Song Dynasty by the inventor Bi Sheng ( 990 -- 1051 ) . Subsequently in 1377 , the world 's oldest extant movable metal print book , Jikji , was printed in Korea during the Goryeo dynasty . Because of this , the diffusion of both movable - type systems was , to some degree , limited to primarily East Asia , although various sporadic reports of movable type technology were brought back to Europe by Christian missionaries , traders and business people who were returning to Europe after having worked in China for several years and influenced the development of printing technology in Europe . Some of these medieval European accounts are still preserved in the library archives of the Vatican and Oxford University among many others . Around 1450 Johannes Gutenberg , introduced the metal movable - type printing press in Europe , along with innovations in casting the type based on a matrix and hand mould . The small number of alphabetic characters needed for European languages was an important factor . Gutenberg was the first to create his type pieces from an alloy of lead , tin , and antimony -- and these materials remained standard for 550 years . \\
     \toprule
     \multicolumn{1}{|c|}{\textbf{Responses}} \\
     \toprule
     \textbf{GPT 3.5} \\
     \midrule
    The invention of the printing press is commonly credited to Johannes Gutenberg, who introduced the metal movable-type printing press in Europe around 1450. Gutenberg's innovations in casting the type based on a matrix and hand mould, along with the use of an alloy of lead, tin, and antimony for creating the type pieces, are considered significant advancements in printing technology.	 \\
    \toprule
    \textbf{Mistral} \\ 
	\midrule 
	 Johannes Gutenberg of Mainz, Germany is given credit for inventing the metal movable - type printing system in Europe and the printing press. \\ 
	\toprule 
    \textbf{\datasetmodel{}} \\
    \midrule
    The world's first movable type printing press technology for printing paper books was invented around AD 1040 in China during the Northern Song Dynasty by the inventor Bi Sheng (990 -- 1051). Around 1450 Johannes Gutenberg introduced the metal movable - type printing press in Europe.		\\
    \toprule
    \textbf{Reference} \\
    \midrule
    The world 's first movable type printing press technology for printing paper books was made of porcelain materials and was invented around AD 1040 in China during the Northern Song Dynasty by the inventor Bi Sheng ( 990 -- 1051 ) . Around 1450 Johannes Gutenberg , introduced the metal movable - type printing press in Europe.		\\
    \bottomrule 
    \end{tabular}
    \caption{The GPT and Mistral answers look good but are only partially faithful to the passage, because they does not give any credit to the innovation in China that is also mentioned in the passage. The GPT answer is also too long.}
    \label{tab:partially_faithful}
\end{table*}

\begin{table*}
\small
    \begin{tabular}{p{15.5cm}}
    \toprule
     \textbf{Question:} who played scotty in the new star trek movie \\
     \toprule
     \textbf{Passage:} Urban was born in Wellington , New Zealand . His father , a German immigrant , owned a leather goods store , and his mother once worked for Film Facilities in Wellington . Through his mother , the young Urban was exposed to classic New Zealand cinema , and developed an interest in the film industry . Urban attended St Mark 's Church School , where he showed an early love for public performance . His first acting role came at age eight , when he had a single line in one episode of the New Zealand television series Pioneer Woman . Though continuing to take part in school stage productions , he did not act professionally again until after high school . \\
     \toprule
     \multicolumn{1}{|c|}{\textbf{Responses}} \\
     \toprule
     \textbf{GPT 3.5} \\
     \midrule
    Simon Pegg played Scotty in the new Star Trek movies.		 \\
    \toprule
    \textbf{\datasetmodel{}} \\
    \midrule
    No answer	\\
    \toprule
    \textbf{Reference} \\
    \midrule
    No answer	\\
    \bottomrule 
    \end{tabular}
    \caption{GPT finds the correct answer but it is not grounded in the document.}
    \label{tab:correct_not_faithful}
\end{table*}

\begin{table*}
\small
    \begin{tabular}{p{15.5cm}}
    \toprule
     \textbf{Question:} where would a commercial quantity of cobalt-60 be produced \\
     \toprule
     \textbf{Passage:} Cobalt - 60 , Co , is a synthetic radioactive isotope of cobalt with a half - life of 5.2714 years . It is produced artificially in nuclear reactors . Deliberate industrial production depends on neutron activation of bulk samples of the monoisotopic and mononuclidic cobalt isotope Co . Measurable quantities are also produced as a by - product of typical nuclear power plant operation and may be detected externally when leaks occur . In the latter case ( in the absence of added cobalt ) the incidentally produced Co is largely the result of multiple stages of neutron activation of iron isotopes in the reactor 's steel structures via the creation of Co precursor . The simplest case of the latter would result from the activation of Fe . Co decays by beta decay to the stable isotope nickel - 60 ( Ni ) . The activated nickel nucleus emits two gamma rays with energies of 1.17 and 1.33 MeV , hence the overall nuclear equation of the reaction is Co  $+ n \rightarrow$ Co $\rightarrow$ Ni $ + e + \nu +$ gamma rays . \\
     \toprule
     \multicolumn{1}{|c|}{\textbf{Responses}} \\
     \toprule
     \textbf{GPT 3.5} \\
     \midrule
    A commercial quantity of cobalt-60 is typically produced through neutron activation of bulk samples of the monoisotopic and mononuclidic cobalt isotope Co. This industrial production process usually takes place in nuclear reactors. Additionally, measurable quantities of cobalt-60 can also be produced as a by-product of typical nuclear power plant operations. However, it's important to note that cobalt-60 is highly regulated and controlled due to its radioactive nature.	 \\
    \toprule
    \textbf{\datasetmodel{}} \\
    \midrule
    Cobalt-60 is produced artificially in nuclear reactors. Measurable quantities are also produced as a by - product of typical nuclear power plant operation.	\\
    \toprule
     \textbf{Mistral} \\ 
	\midrule 
	 Cobalt-60 is typically produced in nuclear reactors through the irradiation of natural cobalt. Commercial quantities are produced by specialized companies in several countries including Argentina, Canada, and Russia, which are the largest suppliers of Cobalt-60 in the world. These companies have dedicated facilities for the production of Cobalt-60 using nuclear reactors. The irradiated cobalt is then extracted and purified to produce the commercial product. \\ 
	\toprule 
    \textbf{Reference} \\
    \midrule
    It is produced artificially in nuclear reactors. Measurable quantities are also produced as a by - product of typical nuclear power plant operation.	 \\
    \bottomrule 
    \end{tabular}
    \caption{In this example the GPT and Mistral response are too long and have both irrelevant and extra information. Mistral is also hallucinating.}
    \label{tab:gpt_too_long}
\end{table*}

\begin{table*}
\small
    \begin{tabular}{p{15.5cm}}
    \toprule
     \textbf{Question:} where are nimbus clouds found in the sky \\
     \toprule
     \textbf{Passage 1 (GOLD):} Nimbostratus has a diffuse cloud base generally found anywhere from near surface in the low levels to about 3,000 m ( 9,800 ft ) in the middle level of the troposphere . Although usually dark at its base , it often appears illuminated from within to a surface observer . Nimbostratus usually has a thickness of about 2000 to 4000 m . Though found worldwide , nimbostratus occurs more commonly in the middle latitudes . It is coded C 2 on the SYNOP report .  \\
     \textbf{Passage 2:} Nimbostratus occurs along a warm front or occluded front where the slowly rising warm air mass creates nimbostratus along with shallower stratus clouds producing less rain , these clouds being preceded by higher - level clouds such as cirrostratus and altostratus . Often , when an altostratus cloud thickens and descends into lower altitudes , it will become nimbostratus . \\
     \textbf{Passage 3:} Multi-level nimbostratus is physically related to other stratiform genus - types by way of being non-convective in nature . However , the other sheet - like clouds usually each occupy only one or two levels at the same time . Stratus clouds are low - level and form from near ground level to 2,000 metres ( 6,600 ft ) at all latitudes . In the middle level are the altostratus clouds that form from 2,000 metres ( 6,600 ft ) to 7,000 metres ( 23,000 ft ) in polar areas , 7,000 metres ( 23,000 ft ) in temperate areas , and 7,600 metres ( 24,900 ft ) in tropical areas ... \\
     \toprule
     \multicolumn{1}{|c|}{\textbf{Responses}} \\
     \toprule
     \textbf{GPT 3.5} \\
     \midrule
    \textbf{Gold} Nimbostratus clouds are generally found anywhere from near the surface in the low levels to about 3,000 m (9,800 ft) in the middle level of the troposphere. They can be found worldwide, but they occur more commonly in the middle latitudes.	 \\
    \toprule
    \textbf{RAG} Nimbostratus clouds are found in the middle level of the troposphere, typically anywhere from near the surface to about 3,000 meters (9,800 feet) in altitude.	 \\
    \toprule
     \textbf{Mistral} \\
     \midrule
    \textbf{Gold} Nimbostratus clouds are found in the middle levels of the troposphere, at altitudes ranging from near the surface up to about 3,000 meters (9,800 ft).	 \\
    \textbf{RAG} 
    Nimbostratus clouds are found in the middle levels of the troposphere, from near surface to about 3,000 m ( 9,800 ft ) in altitude. They are often preceded by higher-level clouds such as cirrostratus and altostratus and are characterized by their thick, dark, and uniform appearance. Nimbostratus clouds are usually associated with frontal systems and are responsible for producing continuous and heavy precipitation.	 \\
    \toprule
    \textbf{\datasetmodel{}} \\
    \midrule
    \textbf{Gold} Nimbostratus has a diffuse cloud base generally found anywhere from near surface in the low levels to about 3,000 m ( 9,800 ft ) in the middle level of the troposphere. Though found worldwide, nimbostratus occurs more commonly in the middle latitudes. \\
   \textbf{RAG}  Nimbostratus has a diffuse cloud base generally found anywhere from near surface in the low levels to about 3,000 m ( 9,800 ft ) in the middle level of the troposphere. Stratus clouds are low - level and form from near ground level to 2,000 metres ( 6,600 ft ) at all latitudes. In the middle level are the altostratus clouds that form from 2,000 metres ( 6,600 ft ) to 7,000 metres ( 23,000 ft ) in polar areas, 7,000 metres ( 23,000 ft ) in temperate areas, and 7,600 metres ( 24,900 ft ) in tropical areas. \\
    \toprule
    \textbf{Reference} \\
    \midrule
    Nimbostratus has a diffuse cloud base generally found anywhere from near surface in the low levels to about 3,000 m in the middle level of the troposphere. Though found worldwide, nimbostratus occurs more commonly in the middle latitudes.	 \\
    \bottomrule 
    \end{tabular}
    \caption{In this example we compare the responses when just the GOLD answer is provided and the top 3 passages are provided when the first passage is the GOLD passage. All models provide good responses when only the gold answer is provided. In the non-RAG setting \datasetmodel{} and Mistral provided irrelevant information from the other passages about other clouds. All three missed the sentence that they can be found world-wide in the RAG setup.}
    \label{tab:gold_in_rag}
\end{table*}

\begin{table*}[t]
\small
    \begin{tabular}{p{15.5cm}}
    \toprule
     \textbf{Question:} who was glumdalclitch how did she help gulliver \\
     \toprule
     \multicolumn{1}{|c|}{\textbf{Responses}} \\
     \toprule
     \textbf{GPT 3.5 Gold} \\
     \midrule
    Glumdalclitch was a nine-year-old girl, the daughter of a giant farmer who captured Gulliver. She became fond of Gulliver and treated him like a pet, creating a miniature bed-chamber for him and playing with him as if he were a doll. Glumdalclitch was also a skilled seamstress and made dolls' clothes for Gulliver. While she provided companionship and amusement for Gulliver, it doesn't mention any specific ways in which she helped him beyond that. \\
    \toprule
     \textbf{Mistral Gold} \\
     \midrule
    Glumdalclitch was a nine-year-old girl from Brobdingnag who captured Gulliver and made him her pet. She was only forty feet tall, making Gulliver appear gigantic in comparison. Despite her small size, she was skilled in sewing and made clothes for Gulliver from the finest Brobdingnagian fabric. However, the fabric irritated Gulliver's skin. Glumdalclitch was amused by Gulliver and enjoyed playing with him as if he were a doll. She gave him the pet name of Glumdalclitch, which means \"little nurse\" in the Brobdingnagian language. Despite being a captive, Gulliver grew very fond of Glumdalclitch and was proud to perform stunts for her entertainment.	 \\
    \toprule
    \textbf{\datasetmodel{} Gold} \\
    \midrule
Glumdalclitch is a skilled seamstress with a talent for making dolls' clothes. She makes Gulliver her pet, creates a small travel case for him, and is amused to play with him as if he were a doll. Gulliver grows very proud of the stunts that he performs for Glumdalclitch's amusement. \\
    \toprule
    \textbf{Reference} \\
    \midrule
    Glumdalclitch is a nine-year-old, forty feet tall daughter of the farmer who captured Gulliver. She made Gulliver her pet, created a bed for him out of a travel case and made him clothes. \\
    \bottomrule 
    \end{tabular}
    \caption{An example that can be used with Table~\ref{tab:gold_not_in_rag} to compare the GOLD setup with the RAG setup when the GOLD passage is not retrieved during RAG.}
    \label{tab:gold_for_RAG}
\end{table*}

\begin{table*}[t]
\small
    \begin{tabular}{p{15.5cm}}
    \toprule
     \textbf{Question:} who was glumdalclitch how did she help gulliver \\
     \toprule
     \textbf{Passage 1:} When the Queen of Brobdingnag takes Gulliver into her court , he has Glumdalclitch brought to court with him . The prideful Gulliver thinks of himself as being greatly honored and promoted by moving to court , but never ceases to love and seek the approval of the little girl who first helped him . Indeed , he remembers her fondly even after returning to England . \\
     \textbf{Passage 2:} Glumdalclitch is the name Gulliver gives his `` nurse '' in Book II of Jonathan Swift 's Gulliver 's Travels . In Book I , Gulliver travels to the land of Lilliput . Leaving there , he travels to the land of Brobdingnag . In Lilliput , Gulliver was a giant , and in Brobdingnag , he is a dwarf , with the proportions reversed . \\
     \textbf{Passage 3:} This article is written like a personal reflection or opinion essay that states a Wikipedia editor 's personal feelings about a topic . Please help improve it by rewriting it in an encyclopedic style . ( October 2017 ) ( Learn how and when to remove this template message ) Gulliver \\
     \toprule
     \multicolumn{1}{|c|}{\textbf{Responses}} \\
    \toprule
    \textbf{GPT 3.5 RAG} \\
    \midrule
    Glumdalclitch was a character in Jonathan Swift's novel Gulliver's Travels. She was the nurse or caretaker of Gulliver in the land of Brobdingnag. Glumdalclitch helped Gulliver by taking care of him, ensuring his safety, and assisting him during his time in the court of the Queen of Brobdingnag. While Gulliver felt honored and promoted in the court, he still cherished and sought approval from Glumdalclitch, showing his deep affection for her. \\
    \toprule
    \textbf{Mistral RAG} \\
    \midrule
    Glumdalclitch was a little girl whom Gulliver met in Brobdingnag, a land of giants, where he was a dwarf. She helped him by nursing him and taking care of him when he was ill. After Gulliver's return to England, he remembered her fondly. \\
    \toprule
    \textbf{\datasetmodel{} RAG} \\
    \midrule
    Glumdalclitch is the name Gulliver gives his nurse'' in Book II of Jonathan Swift's Gulliver's Travels. In Lilliput, Gulliver was a giant, and in Brobdingnag, he is a dwarf, with the proportions reversed. \\
    \toprule
    \textbf{Reference} \\
    \midrule
    Glumdalclitch is a nine-year-old, forty feet tall daughter of the farmer who captured Gulliver. She made Gulliver her pet, created a bed for him out of a travel case and made him clothes. \\
    \bottomrule 
    \end{tabular}
    \caption{Alternative appropriate and relevant answers based on the retrieved passages compared to the GOLD setup in Table~\ref{tab:gold_for_RAG}.}
    \label{tab:gold_not_in_rag}
\end{table*}



\begin{table*}
\small
	\begin{tabular}{p{15.5cm}}
	\toprule
	\multicolumn{1}{|c|}{\textbf{Conversation}} \\ 
	\toprule 
	\textbf{User}: why did the us demand trade with japan \\ 
	\toprule 
	\multicolumn{1}{|c|}{\textbf{Passages}} \\ 
	\toprule 
	\textbf{Passage 1} \\ 
	\midrule 
	The United States reacted by seeking to bring the Japanese war effort to a complete halt by imposing a full embargo on all trade between the United States to Japan on 1 August 1941 , demanding that Japan withdraw all troops from both China and Indochina . Japan was dependent on the United States for 80 percent of its oil , resulting in an economic and military crisis for Japan that could not continue its war effort with China without access to petroleum and oil products . Attack \\ 
	\toprule 
	\textbf{Passage 2} \\ 
	\midrule 
	The U.S. embargoes gave Japan a sense of urgency . It would either have to agree to Washington 's demands or use force to gain access to the resources it needed . \\ 
	\toprule 
	\textbf{Passage 3} \\ 
	\midrule 
	Japan 's goal after 1931 was economic dominance of most of East Asia , often expressed in Pan-Asian terms of `` Asia for the Asians . '' . Japan was determined to dominate the China market , which the U.S. and other European powers had been dominating . On October 19 , 1939 , the American Ambassador to Japan , Joseph C. Grew , in a formal address to the America - Japan Society stated : \\ 
	\toprule 
	\multicolumn{1}{|c|}{\textbf{Responses}} \\ 
	\toprule 
	\textbf{GPT 3.5 RAG} \\ 
	\midrule 
	The United States demanded a halt to trade with Japan in response to Japan's aggressive actions in East Asia and its invasion of China. The U.S. sought to put pressure on Japan to withdraw its troops from China and Indochina by imposing an embargo on trade. Additionally, the U.S. was concerned about Japan's growing economic dominance in the region, particularly in the China market, which had previously been dominated by the U.S. and other European powers. By cutting off trade, the U.S. hoped to weaken Japan's war effort and force a resolution to the conflict through diplomatic means. \\ 
	\toprule 
	\textbf{Mistral RAG} \\ 
	\midrule 
	 The United States demanded trade with Japan not because it wanted to help Japan, but because it wanted to ensure its own economic interests and prevent Japan from dominating the China market and expanding its military power in the Pacific. The U.S. embargo on oil and other resources was a response to Japan's aggression in China and its refusal to withdraw its troops from that region. This economic pressure forced Japan into a corner and ultimately led to its decision to attack the U.S. naval base at Pearl Harbor on December 7, 1941. The attack was a desperate move by Japan to secure the resources it needed to continue its war efforts. \\ 
	\toprule 
	\textbf{\datasetmodel{} RAG} \\ 
	\midrule 
	Japan was dependent on the United States for 80 percent of its oil, resulting in an economic and military crisis for Japan that could not continue its war effort with China without access to petroleum and oil products. \\ 
	\bottomrule 
     \caption{Example of the RAG setting where all retrieved passages are irrelevant. The question asks about US demanding trade with Japan, while the passages talk about the reverse situation - US stopping (embargo) trade with Japan or Japan wanting trade.}
    \label{tab:bad_retrieval}
	\end{tabular}
\end{table*}

\end{document}